\documentclass[runningheads]{llncs}
\usepackage[T1]{fontenc}
\usepackage{graphicx}
\usepackage{hyperref} 

\usepackage{booktabs}

\usepackage{todonotes}

\makeatletter
\newcommand{\printfnsymbol}[1]{%
  \textsuperscript{\@fnsymbol{#1}}%
}
\makeatother

\usepackage{color}

\begin{document}
\title{ACCSAMS: Automatic Conversion of Exam Documents to Accessible Learning Material for Blind and Visually Impaired}
\titlerunning{ACCSAMS}

\author{David Wilkening\inst{1}\orcidID{0009-0000-2516-1706}
\and
Omar Moured\inst{1,2}\orcidID{0000-0003-4227-8417}
\and
Thorsten Schwarz\inst{1}\orcidID{0000-0002-7346-5744}
\and
Karin Müller\inst{1}\orcidID{0000-0003-4309-1822}
\and
Rainer Stiefelhagen\inst{1,2}\orcidID{0000-0001-8046-4945}
}
\authorrunning{D. Wilkening et al.}

\institute{
ACCESS@KIT, Karlsruhe Institute of Technology, Germany\\
\and
CV:HCI@KIT, Karlsruhe Institute of Technology, Germany\\
\email{mail@david-wilkening.de}, \email{moured.omar@gmail.com}\\
\url{https://gitlab.access.kit.edu/David/accsams/}
}

\maketitle
\begin{abstract}
Exam documents are essential educational materials for exam preparation. However, they pose a significant academic barrier for blind and visually impaired students, as they are often created without accessibility considerations. Typically, these documents are incompatible with screen readers, contain excessive white space, and lack alternative text for visual elements. This situation frequently requires intervention by experienced sighted individuals to modify the format and content for accessibility. We propose ACCSAMS, a semi-automatic system designed to enhance the accessibility of exam documents. Our system offers three key contributions: (1) creating an accessible layout and removing unnecessary white space, (2) adding navigational structures, and (3) incorporating alternative text for visual elements that were previously missing. Additionally, we present the first multilingual manually annotated dataset, comprising 1,293 German and 900 English exam documents which could serve as a good training source for deep learning models.

\end{abstract}

\section{Introduction}
The availability of accessible course material is often a critical barrier for students with blindness and visual impairment (BVI)\cite{reed_2012}. An important type of learning material is past exams, which give students the opportunity to understand the level of knowledge expected in an exam and practice their acquired knowledge\cite{van_etten_1997}. However, it is often not possible for BVI students to directly access these exam documents, often requiring sighted individuals to carefully modify the format and content to ensure compatibility with screen readers.

Documents are accessed primarily by BVI individuals using various assistive technologies, depending on their preference and availability \cite{korey_2020}. Such documents need to follow guidelines\cite{wcag} to ensure accessibility. However, \textbf{the majority of exams are not created as part of an ongoing course in which BVI students participate. Therefore, they are usually created without considering accessibility.} To address these challenges, we developed a semi-automatic AI-based method that aids in converting exam materials into an accessible format. This involves identifying key content, solutions, adding navigational structures, and including missing alternative texts. Our contributions include: (1) Identifying a preferred exam format for BVI students through interviews, (2) Developing a system for efficient document conversion, and (3) Creating the first multilingual dataset of 1,293 German and 900 English exams.

\section{Related Work}
Making already existing documents accessible after their creation is ongoing research with many subtasks. \textit{SciA11y} \cite{scia11y_2021} is a system specifically developed to convert \textit{PDFs} of scientific papers into accessible \textit{HTML} formats. It transforms multicolumn documents into a linear, one-dimensional layout that screen readers can easily process by creating \textit{HTML} header tags for each section title. Furthermore, \textit{SciA11y} generates accessible hyperlinks for citations and references, facilitating seamless navigation for BVI users. \textit{Nougat} \cite{blecher2023nougat} is a transformer encoder-decoder model that converts document images into markdowns in an end-to-end manner. Contrary to \textit{SciA11y} it is also capable of transforming math equations and simple tables. However, (1) both are only tailored for scientific publications and (2) do not address the challenges associated with providing alt-text for all types of content blocks.

Inspired by \textit{SciA11y}, we built a pipeline to convert existing documents into an accessible format. We made modifications to take into account the domain specifics of exam documents and their target use as learning materials. We also address some shortcomings, such as support for mathematical equations in text.
\section{Need-finding Interviews}

The adaptation of exam materials into different formats is not just for accessibility but also practiced by KIT's departmental student representatives. A survey\footnote{\url{https://www.fsmi.uni-karlsruhe.de/Fachschaft/Umfrage/}} among KIT's math and computer science students has revealed that 79.9\% of the respondents prefer a reconfiguration of document layouts to eliminate superfluous white space, thus reducing both the cost and paper consumption associated with printing (n = 369). 76.8\% of the participants prefer to relocate the sets of solutions from their original placement within the questions to a consolidated section at the end of the document (n = 362).

We conducted interviews to identify how students with BVI use exam documents as learning materials. These interviews were instrumental in determining the type of semantic information that ACCSAMS needs to extract from document files and how the results should be presented to serve as effective learning resources for students with BVI. They served to expand on our knowledge from our experience as a Disability Support Office (DSO) in making documents accessible and remixing exam documents for students with vision in the departmental student representations.

\subsection{Participants}
The interviews were conducted with two blind male graduate students in computer science and physics. Both make use of the document conversion service for BVI at KIT.

\subsection{Procedure}
We asked participants the following three questions: (1) Which exam format do you prefer (e.g., \textit{ PDF, HTML, WORD}, Markdown) and why? (2) How do you typically interact with exam documents for the purpose of exam preparation? By annotating a separate document or directly editing the file? (3) What layout find most effective for navigating the questions and answers? Each question followed immediately by its answer, or all questions first followed by all answers?


\subsection{Findings} 
Both participants preferred inline editing formats like Word and Markdown over tagged PDFs. They reported difficulties with navigation due to the lack of screen reader-friendly hierarchy in exam documents, which limits quick movement between questions and answers, and noted the absence of alt-text for figures and equations. Regarding layout, opinions differed: one favoured separate documents for solutions, while the other preferred solutions under each question. In response, we outlined design steps to enhance exam accessibility in our UI, including (1) analyzing document layout to identify content blocks and solutions, (2) extracting document hierarchy to support screen reader navigation, (3) adding alt-text to visual elements, and (4) ending with the option to export the document in either Markdown or Word format with the option to provide alternative reading orders by repositioning solutions.

\section{Creation of the Exam Documents Dataset}
We collected exam materials from the Common Crawl project\footnote{\url{https://commoncrawl.org/}}, a public archive of web data. We analyzed eight datasets, released between mid 2022 and the end of 2023, extracting \textit{PDFs}, which include ``exam'' or ``klausur'' as a keyword in the URL. Positive matches were downloaded. The text layer of the locally available \textit{PDFs} was utilized to refine the data set by iteratively developing a list of common false positive keywords (such as ``exam schedule''). In addition, a keyword search was used to suggest a study field for each document.

Subsequently, the remaining documents were manually filtered. Furthermore, we identified whether an exam originated from higher education (university, college or similar) or another source. Unique types of exams (such as entrance exams) were included if they shared a similar format as other exams. As a result, we obtained a collection of 1,293 German (361 computer science, 337 physics, 216 mathematics, 175 economics, 67 law, 33 electrical engineering, 32 mechanical engineering) and 900 English exams (341 computer science, 239 chemistry, 150 math, 124 economics) in native \textit{PDF} format, originating from 281 different domains.

\section{The ACCSAMS System}

\begin{figure}
  \centering
  \includegraphics[width=\textwidth]{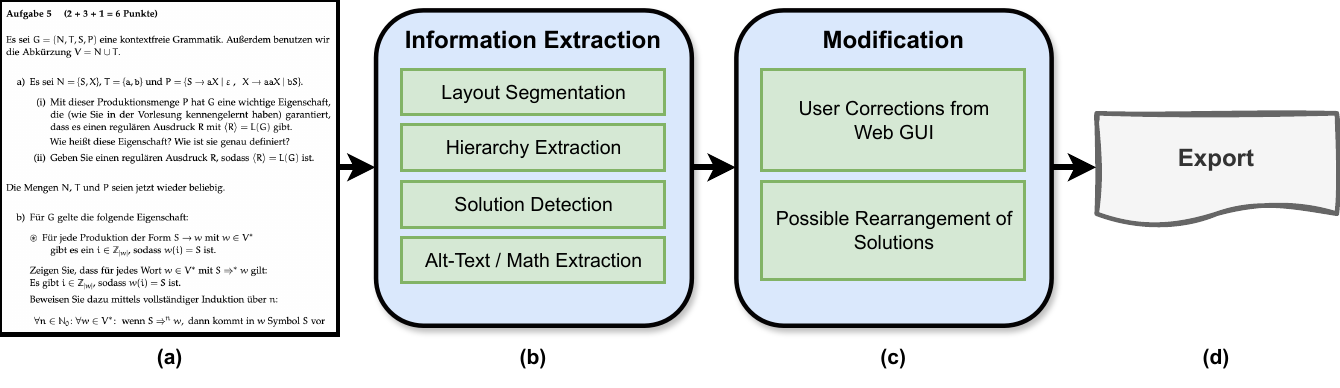}
  \label{fig:accsams}
  \caption{The ACCSAMS pipeline: (a) Exam page. (b) Automated processing (c) User application of corrections and repositioning of solutions (d) Output to accessible markdown or conversion to WORD format.}
\end{figure}

The system is based on 3 steps: (1) segmenting the layout of every page, extracting all relevant content blocks, while establishing if they are specific to a solution, (2) determining the reading order and hierarchy between the content blocks, and (3) generating alt-texts automatically or requesting them from the user for figures and tables through our Web GUI.

\subsection{Layout segmentation}

The layout segmentation module analyzes the document and splits it into content blocks using YOLOv8\footnote{\url{https://github.com/ultralytics/ultralytics}} as the AI detection model. We first pre-trained it on DocLayNet\cite{doclaynet} a dataset of a diverse collection of 80,863 pages from various domains. 
We then fine-tune it on our own exam dataset. For this, we annotated 1917 pages and split them into a training and evaluation corpus (80/20 ratio). The classes and the model results are in Table \ref{tab:detection_metrics}.

\subsection{Hierarchy Extraction}
To enhance navigation accessibility for screen reader users, extracting and organizing the document's hierarchy is essential. We achieve this by structuring content blocks into a coherent tree structure, applying two heuristic rules. The first rule involves sorting content blocks by their location within the document, using page number and vertical and horizontal coordinates as sorting criteria. The second rule focuses on categorizing list items and headings. This categorization is based on the content's textual characteristics and enumeration styles, such as Roman, numeric, alphabetic, or heading levels. By implementing these strategies, we construct a tree structure that accurately represents the hierarchy of the document.

\subsection{Solution detection}
In order to facilitate the relocation of solutions to different parts of a document or to insert hints for screen readers, it is necessary to identify the solutions accordingly. To achieve this, we follow two rules. First, if a content block is a heading and includes keywords such as ``solution'' or ``answer'', both the heading itself and all its descendants in the subtree are marked as solution specific content blocks. Our layout segmentation model is trained to not incorrectly identify text elements such as ``your answer: \underline{\hspace{1cm}}'' as headings. Second, if a content block contains color, it is also identified as a solution.

\subsection{User Interface}
The web application was implemented to assist users in individual steps to enhance the accessibility of exam documents. These steps include: (1) Upload an exam, (2-5) verify the extracted content blocks, question / solution assignment, alt text, and hierarchy identified by the pipeline, and finally, (6) export the exam to the desired layout and file format. 



\section{Evaluation}
To evaluate the detection of relevant content blocks, we have annotated 1917 pages from our exam dataset and split them into a training (1,533) and evaluation (384 pages) corpus (80/20 ratio).
The results can be seen in \autoref{tab:detection_metrics}.

\subsection{System Evaluation}
For reading order and hierarchy extraction, we manually annotated 26 different documents originating from different domains to ensure a variety of authors and styles.
The evaluation of the reading order was done using the Average Relative Distance (ARD) metric according to \cite{layoutreader}. We achieve an average AVD (per complete document) of 6.23 with a standard deviation of 10.05. The hierarchy is evaluated on the same 26 documents by comparing the hierarchy level of each content block. We measure this with an average Euclidean distance of 0.25 (standard deviation 0.43) and for their relative (between two consecutive content blocks in reading order) distance of 0.09 (standard deviation 0.08).

\begin{table}
    \centering
\begin{footnotesize}
        \caption{Detection model performance evaluated on 384 test pages.}
    \begin{tabular}{lccccc}
    \toprule
    Classes & \# Instances & Precision & Recall & mAP50 & mAP50-95 \\
    \midrule
    Headings & 446 & 0.884 & 0.948 & 0.956 & 0.86 \\
    Paragraphs & 3475 & 0.944 & 0.919 & 0.965 & 0.875 \\ 
    List symbols & 1933 & 0.99 & 0.936 & 0.987 & 0.747 \\ 
    Figures & 246 & 0.815 & 0.878 & 0.898 & 0.814 \\
    Formulas & 271 & 0.8 & 0.764 & 0.848 & 0.774 \\
    Tables & 75 & 0.811 & 0.88 & 0.878 & 0.852 \\
    All & 6446 & \textbf{0.874} & \textbf{0.887} & \textbf{0.922} & \textbf{0.821} \\
    \bottomrule
    \end{tabular} 
    \label{tab:detection_metrics}
\end{footnotesize}
\end{table}

\subsection{User Study of User Interface}

To evaluate the end-user web application of ACCSAMS, we successfully enlisted 4 people with prior experience in transforming examination documents into study resources. Two of these participants are employed at the ACCESS@KIT literature conversion service. The remaining two are members of the joint departmental student representation for the mathematics and computer science faculties.

\subsubsection{Procedure}

Upon completion of the consent form and the provision of demographic data, the participants were invited to participate in individual online think-aloud sessions. Using their own personal devices, they participated in the sessions using the Firefox or Chrome web browser. An elucidation of the purpose of the application, accompanied by a demonstration of the system, was presented. Subsequently, the participants received a simple two-page math examination document with questions as an initial warm-up exercise. This was followed by a physics examination, comprising 19 non-empty pages, with different documents of questions and solutions. The session culminated with the opportunity for participants to engage in open-ended questions, allowing them to reflect on their experiences and draw parallels with existing tools with which they are familiar for document conversion. In conclusion, participants received a System Usability Scale (SUS) questionnaire.

\subsubsection{Results}

The system received an average SUS score of 84. All participants agreed that the system is an improvement over the tools they are currently using for document conversion.

All participants expressed the wish for immediate feedback on the impact of their actions. One participant requested a comprehensive view of all subtask predictions, such as layout segmentation, solution detection, hierarchy extraction, and alt texts, with the option to correct inaccuracies in the specific sub task editor. A trio of users shared a common observation that a meticulous review of the entire exam content is required at each stage to validate the predictions, an exercise that rendered the process somewhat daunting and repetitive.

Two participants were surprised by the need to manually input metadata such as the title, duration, and date of the exam. They felt that grading points are significant in layout segmentation. One participant typically places these points in the task's heading, aiding screen reader users in navigating the table of contents. Both participants were unsure about how to handle points for specific anticipated keywords in a student's response. ACCSAMS does not dictate how to handle this, allowing users to incorporate points into \texttt{heading} or \texttt{text} content blocks as they see fit. One user chose not to label these blocks, excluding them from the output.
\section{Discussion and Limitations}
\textbf{Intelligent Features:} Using predictive algorithms to divide document conversion into subtasks reduces active editing and simplifies the conversion process. However, this method requires a thorough review of the document for each subtask. Each subtask editor increases cognitive load, leading to user fatigue. Our application's framework mirrors the ACCSAMS pipeline's steps. Improving the repetitive ``review and correct'' process in the user interface could enhance user satisfaction in future versions.

\textbf{Handling Grading Points} Grading points were not seen as special elements of examination documents. Their information is seamlessly integrated within the concept of \texttt{heading} and \texttt{text} content blocks without the need for specialized processing. However, users identified these as distinct elements and experienced uncertainty in dealing with them, given the lack of specialized recognition in the user interface.

\textbf{Specific Question Formats} ACCSAMS does not explicitly cater to question formats such as \textit{multiple choice} or \textit{fill in the blanks}. Although the generic nature of ACCSAMS and its user interface allows the conversion of these documents, a more structured and automated methodology would be advantageous to guide users.

\textbf{Need-finding User Study} Our initial investigations, although encouraging, were conducted with a limited number of users and lacked comprehensive control measures. Future studies should involve a larger and more diverse group of BVI individuals. Particularly in light of representing grading points, and dealing with specific question formats such as multiple choice and fill-in texts, there is a need for additional research besides applying generic guidelines for document accessibility.
\section{Conclusion}
In this work, we introduce ACCSAMS, a pipeline and user interface designed to facilitate efficient transfer of exams to accessible learning materials for individuals with or without BVI. By leveraging AI models along with heuristic rules, we empower sighted individuals to seamlessly enhance the accessibility of exam documents. We believe that this initiative is a first step to encourage other researchers to explore additional features aimed at improving the conversion process.
Future work involves further integration of automatic alt-text generation for figures and tables in the pipeline, as well as specialized alt-text editors into the GUI for the various content blocks.

\begin{credits}
\subsubsection{\ackname}
The authors thank Viola Buck Cabrera and Iva Andreeva for the implementation of the end-user web application. In addition, we thank the HoreKa computing cluster at KIT for the computing resources used to conduct this research.

\subsubsection{\discintname}
This research was funded by the European Union’s Horizon $2020$ research and innovation program under the Marie Sklodowska-Curie grant agreements no. $861166$. 

\end{credits}

\bibliographystyle{splncs04}
\bibliography{mybib}
\end{document}